\documentclass[10pt,twocolumn,letterpaper]{article}

\usepackage[pagenumbers]{cvpr} 

\definecolor{cvprblue}{rgb}{0.21,0.49,0.74}
\usepackage[pagebackref,breaklinks,colorlinks,allcolors=cvprblue]{hyperref}

\newcommand{\comment}[1]{} 

\usepackage{booktabs}       
\usepackage{amsfonts}       
\usepackage{nicefrac}       
\usepackage{microtype}      
\usepackage[dvipsnames]{xcolor}         
\usepackage{svg}
\usepackage[skip=2pt]{caption}
\usepackage{makecell}
\usepackage{tikz,pgfplots}
\usetikzlibrary{calc}
\usetikzlibrary{spy}
\usepackage{tabularx}
\usepackage{multirow}
\usepackage{subcaption}
\usepackage{float}

\usepackage{graphicx}
\usepackage{amsmath}
\usepackage{amssymb}
\usepackage{amsthm}

\usepackage{lineno}

\usepackage{pifont}
\newcommand{\xmark}{\ding{55}}%
\newcommand{\gcmark}{\textcolor{Green}{\ding{51}}}
\newcommand{\rxmark}{\textcolor{Red}{\ding{55}}}

\newcommand{\vt}[1]{\mathbf{#1}}
\newcommand{\surf}[1]{\mathcal{S}_{#1}}
\newcommand{\V}[1]{\mathcal{V}(#1)}
\newcommand{\I}{\mathcal{I}}
\renewcommand{\L}{\mathcal{L}}
\renewcommand{\P}{\mathcal{P}}
\newcommand{\dd}{\mathrm{d}}
\newcommand{\norm}[1]{\left\lVert#1\right\rVert}
\newcommand{\dotp}[2]{\langle#1\,,#2\rangle}
\newcommand{\dom}{\Omega}
\newcommand \eq[1]{\begin{equation}\begin{aligned}#1\end{aligned}\end{equation}}
\DeclareMathOperator{\tr}{Tr}
\newcommand{\SAstd}{$\text{SA}\sigma$}

\newcommand*{\inparagraph}[1]{\noindent\textbf{#1}\hspace{0.5em}}

\def\paperID{112} 
\def\confName{CVPR}
\def\confYear{2025}

\title{4Deform: Neural Surface Deformation for Robust Shape Interpolation}

\author{Lu Sang$^{1,2}$, Zehranaz Canfes$^{1}$, Dongliang Cao$^{3}$,\\
Riccardo Marin$^{1,2}$, Florian Bernard$^{3}$, Daniel Cremers$^{1,2}$ \\
$^{1}$Technical University of Munich, $^{2}$Munich Center of Machine Learning\\
{\tt\small \{lu.sang, zehranaz.canfes, riccardo.marin, cremers\}@tum.de} \\
$^{3}$University of Bonn \\
{\tt\small \{dcao, fb\}@uni-bonn.de}
}

\begin{document}
\twocolumn[{
    \renewcommand \twocolumn[1][]{#1}
    \maketitle
    \centering
    \vspace{-0.8cm}
    \includegraphics[width=0.95\textwidth]{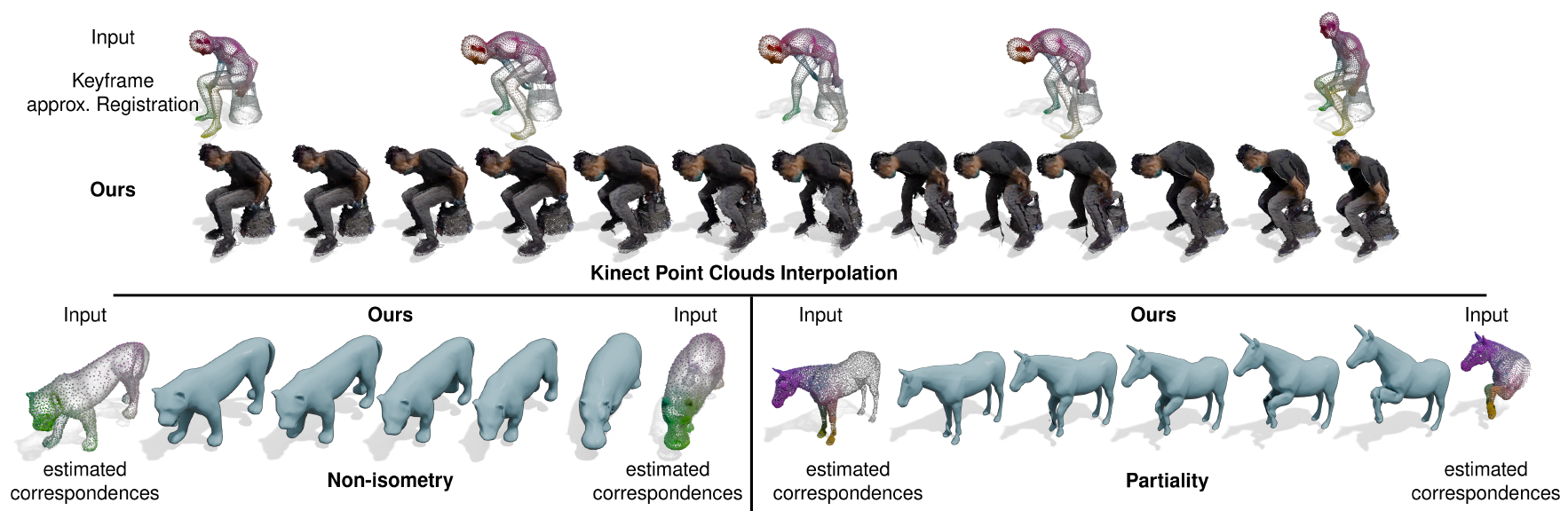}
    \captionof{figure}{
    \textbf{4Deform} takes a sparse temporal sequence of point clouds as input
    and generates realistic intermediate shapes. Starting from just pairs of point clouds and estimated sparse, noisy correspondences (indicated using colors in the point clouds),
    our method obtains realistic long-range interpolations, even for shapes with changing topology (\eg, the human-object interaction in the top row), and can generalize the interpolation results to real-world data (Kinect point clouds in the second row). Meanwhile, our method can handle non-isometrically deformed shapes (bottom left)  as well as partial shapes (bottom right).
    }
    \label{fig:teaser}
    \vspace{1em} 
}]

\begin{abstract}
Generating realistic intermediate shapes between non-rigidly deformed shapes is a challenging task in computer vision, especially with unstructured data (\eg, point clouds) where temporal consistency across frames is lacking, and topologies are changing. Most interpolation methods are designed for structured data (\ie, meshes) and do not apply to real-world point clouds. In contrast, our approach, 4Deform, leverages neural implicit representation (NIR) to enable free topology changing shape deformation. Unlike previous mesh-based methods that learn vertex-based deformation fields, our method learns a continuous velocity field in Euclidean space. Thus, it is suitable for less structured data such as point clouds.
Additionally, our method does not require intermediate-shape supervision during training; instead, we incorporate physical and geometrical constraints to regularize the velocity field. We reconstruct intermediate surfaces using a modified level-set equation, directly linking our NIR with the velocity field. Experiments show that our method significantly outperforms previous NIR approaches across various scenarios (\eg, noisy, partial, topology-changing, non-isometric shapes) and, for the first time, enables new applications like 4D Kinect sequence upsampling and real-world high-resolution mesh deformation.
\vspace*{-12pt}
\end{abstract}
\vspace*{-12pt}    
\section{Introduction}
\label{sec:intro}
Inferring the dynamic 3D world from just a sparse set of discrete observations is among the fundamental goals of computer vision. These observations might come, for example, from video sequences~\cite{sun2021neucon}, Lidar scans~\cite{forlani2006c, tachella2019bayesian} or RGB-D cameras~\cite{Sommer2022, sang2023high}.
Even more challenging is the recovery of plausible motion in between such observations. Despite the relevance of this problem, just a few works addressed this, likely because the solution requires merging concepts of camera-based reconstruction with techniques of 3D shape analysis and interpolation.


In the computer graphics literature, researchers have developed interpolation approaches for mesh representations~\cite{sorkine2007rigid, eisenberger2021neuromorph,baek2015isometric}. 
These often require a dense, exact point-to-point correspondence between respective frames~\cite{cao2024motion2vecsets, cao2024spectral, eisenberger2021neuromorph}, which is usually unpractical and rare in real-world applications. Also, such methods rely on a predefined topology and do not support changes (\eg, partiality, the interaction of multiple independent components). Moreover, the recovered interpolation is defined only on the mesh surface, which limits the applicability to other data forms.


The recent advent of neural implicit fields~\cite{gropp2020implicit, sang2023weight, sitzmann2019siren, haerenstam2024diffcd} opened the door to more flexible solutions. These methods convert the start and final meshes or point clouds into implicit representations and recover the intermediate frames by either latent space optimization~\cite{Novello2023neural, liu2022learning} or deformation modeling~\cite{anonymous2024implicit, yang2021geometry}. 
The theoretical advantage comes from the topological flexibility of implicit representations~\cite{sdf}. However, the latent space-based methods generally do not consider the physical properties of the recovered intermediate shapes and therefore fail to produce reasonable interpolations such as rigid movements~\cite{Novello2023neural, liu2022learning}. 
Some other methods propose physical constraints during optimization~\cite{anonymous2024implicit, yang2021geometry} but they either assume ground truth correspondences~\cite{anonymous2024implicit}
or user-defined handle points~\cite{yang2021geometry}, which makes it fail on complicated deformation or non-isometric deformation.~\cref{tab:summary} summarizes the strengths and limitations of different methods. 
\begin{table}[t]
\footnotesize
\addtolength{\tabcolsep}{-2pt}
\begin{tabular}{lcccccc}
\specialrule{.1em}{.05em}{.05em} 
        \toprule
        Methods & Corr.Est. & Non-iso. & Part. &  Topo. & Seq. &  Real. \\
        \midrule
        SMS~\cite{cao2024spectral} & \gcmark & \gcmark & \rxmark & \rxmark & \gcmark & \rxmark \\
        Neuromorph~\cite{eisenberger2021neuromorph} & \gcmark & \gcmark & \rxmark & \rxmark & \gcmark & \rxmark \\
        LIMP~\cite{Cosmo2020}  & \rxmark & \gcmark & \rxmark & \rxmark & \gcmark & \rxmark \\
        NFGP~\cite{yang2021geometry}  & \rxmark & \rxmark & \rxmark & \gcmark & \rxmark & \rxmark \\
        NISE~\cite{Novello2023neural} & - & \rxmark & \rxmark & \gcmark & \rxmark & \rxmark \\
        LipMLP~\cite{liu2022learning} & - & \rxmark & \rxmark & \gcmark & \gcmark & \rxmark \\
        \cite{anonymous2024implicit} & \rxmark & \gcmark & \gcmark & \gcmark & \rxmark & \gcmark \\
        \textbf{Ours} & \gcmark & \gcmark & \gcmark & \gcmark & \gcmark &\gcmark \\
        \bottomrule
\end{tabular}
\caption{\textbf{Summary of Methods Capability.} We list the capabilities of previous mesh and NIR-based methods. The column Corr. Est. indicates if the method can estimate the correspondences (\gcmark) or needs ground-truth correspondences as input (\rxmark); 
If the methods can handle non-isometric deformation (Non-iso.), partial shape deformation (Part.), topological changes (Topo.), work for Sequences (Seq.), and for the real-world data (Real.).
}
\vspace*{-2mm}
\label{tab:summary}
\end{table}

In this work, we address the challenging task of recovering motion between sparse keyframe shapes, relying only on coarse and incomplete correspondences. Our method begins by establishing correspondences through a matching module, followed by representing the shapes with an implicit field and modeling deformation via a velocity field. We introduce two novel loss functions to minimize distortion and stretching, ensuring physically plausible deformations. Our approach encodes shapes in a latent space, enabling both sequence representation and extrapolation of new dynamics. 

For the first time, we present results that begin with imprecise correspondences obtained from standard shape matching and registration pipelines and even not always spatially aligned with the input points \eg, real-world data. Our method not only outperforms state-of-the-art approaches, even when they are designed to overfit specific frame pairs, but also demonstrates versatility in real-world applications, including Kinect point cloud interpolation for human-motion interaction sequences, upsampling of real captures with partial supervision, and sequence extrapolation, as shown in~\cref{fig:teaser}.
In summary, our contributions are:
\begin{itemize}
    \item A data-driven framework for 3D shape interpolation that merely requires an estimated (noisy and incomplete) correspondence, and for the first time demonstrates applicability to real data such as noisy and partial point clouds.
    \item The derivation of two losses to prevent distortion and stretching in implicit representation, promoting desirable physical properties on the interpolated sequence.
    \item Experimental results confirming state-of-the-art performance on shape interpolation and the applicability to challenging downstream tasks like temporal super-resolution and action extrapolation.
\end{itemize}
Our code will be released for future research.

\section{Related Work}
\label{sec:related}
4D reconstruction from discrete observations involves recovering a continuous deformation space that not only aligns with the input data at specific time steps but also provides plausible intermediate reconstructions. The reconstructed sequences should preserve the input information while filling in any missing details absent in the original data, ultimately creating a complete and coherent representation of the observed sequence. This task involves shape deformation and interpolation.

\subsection{Surface Deformation}
Modeling the 3D dynamic world involves surface deformation, essential in fields like gaming, simulation, and reconstruction. Physically plausible deformations are often needed. However, the task relies heavily on the representation of the data.

\inparagraph{Mesh Deformation.} Mesh deformation typically involves directly adjusting vertex positions within a mesh pair, taking advantage of the inherent neighboring information. This allows mesh-based methods to incorporate physical constraints, such as As-Rigid-As-Possible (ARAP)~\cite{sorkine2007rigid, botsch2006primo}, area-preserving or elasticity loss~\cite{bastianxie2024hybrid} to constrain the deformation. While mesh deformation is well-studied~\cite {cao2023revisiting, cao2024motion2vecsets,eisenberger2018divergencefree, eisenberger2020hamiltonian, eisenberger2021neuromorph}, it requires predefined spatial discretization and fixed vertex connectivity to maintain topology. This leads to challenges with topology changes~\cite{cao2024spectral} and resolution limitations, as methods must process all vertices together. Consequently, techniques like LIMP~\cite{Cosmo2020} downsample meshes to 2,500 vertices, and others~\cite{cao2024spectral, roetzer2024discomatch, eisenberger2021neuromorph} re-mesh inputs to 5,000 vertices, with output resolution tied to these constraints.

\inparagraph{Implicit Field Deformation.} Unlike mesh representations, implicit surface representations offer several advantages. First, neural implicit fields allow flexible spatial resolution during inference, as discretization isn’t pre-fixed. Second, they can represent arbitrary topologies, making them well-suited for complex cases that mesh-based methods struggle with, such as noisy or partial observations. However, directly deforming an implicitly represented surface is challenging because surface properties aren’t explicitly stored, limiting direct manipulation of surface points. This area remains underexplored, with only a few approaches addressing it. For instance, NFGP~\cite{yang2021geometry} introduces a deformation field on the top of an implicit field, constraining it physically using user-defined handle points to match target shapes. This pioneering work directly deforms the implicit field; however, its practical applications are limited as it only provides the starting and ending shapes, with processing times for a single shape pair extending over several hours. Other methods, such as those in~\cite{Novello2023neural, mehta2022level}, focus on shape smoothing or morphing without targeting specific shapes. The work~\cite{anonymous2024implicit} introduced a fast, flexible framework for directly deforming the implicit field, capable of generating physically plausible intermediate shapes. However, as an optimization-based approach, it requires training separately for each shape pair and struggles with handling large deformations.

\subsection{Shape Interpolation}
Shape interpolation is a specialized subset of shape deformation that involves generating intermediate shapes between a start and target shape. The interpolated sequence should reflect a physically meaningful progression,
making it essential that the interpolated shapes are not only geometrically accurate but also serve to complete the narrative implied by the initial and final shapes. Therefore, we emphasize the concept of \textbf{physically plausible} shape interpolation, which should be a guiding principle for tasks in this area. 
There are two main approaches: generative models and physics-based methods. Generative models~\cite{gan, diffusion} rely on extensive datasets to produce shapes, but their outputs can lack realism due to data dependency. Physics-based methods, like~\cite{Novello2023neural, liu2022learning, anonymous2024implicit}, optimize over a shape pair to generate intermediates but may have limited applications.
Another way is to learn a deformation space of shapes, such as using latent space~\cite{Goodfellow2016}, and hope that generating intermediate shapes is equivalent to interpolating the latent shape~\cite{Cosmo2020}. However, such methods suffer from the same problem as generative models in that the latent space interpolation may be far from physically plausible.\\
To address these limitations, we propose a lightweight solution that can be trained on datasets of any size. Our model adopts an AutoDecoder architecture to maintain generative capability and combines physical and geometric constraints to ensure the generated results are physically plausible. 



\section{From Implicit Surface Deformation ...}

\label{sec:background}
\inparagraph{Implicit representation of the moving surface.}
We represent the moving surface $\surf{t}$ in the volume $\dom\subset\mathbb{R}^3$ implicitly as the zero-crossing of a time-evolving signed distance function $\phi:\dom\times [0,T]\rightarrow\mathbb{R}$:
\eq{\label{eq:time_implicit}
\surf{t} = \{ \vt{x} \in \dom | ~ \phi(\vt{x},t) = 0\}\;.
}
This implies that
\eq{\label{eq:time_implicit}
\phi(\surf{t},t)=0 \quad\forall t.
}
It follows that the total time derivative is zero, i.e.
\eq{\label{eq:les}
\frac{\dd}{\dd t}\phi(\vt{x},t) = \partial _t \phi + \mathcal{V}^\top \nabla\phi   = 0,
}
where $\mathcal{V}=\frac{d}{dt}\surf{t}$ denotes the velocity of the moving surface.  Eq.~\eqref{eq:les} is known as the {\em level-set equation}~\cite{Dervieux-Thomasset-79,Dervieux-Thomasset-81}.  It tells us how to move the implicitly represented surface $\surf{t}$ along the vector field $\mathcal V$ by deforming the level-set function $\phi$. Since over time, the level-set function will generally lose its property of being a signed distance function, we add an Eikonal regularizer with a weight $\lambda_l$ to obtain the modified level-set equation~\cite{anonymous2024implicit}, i.e.
\eq{\label{eq:lse3}
   \partial _t \phi + \mathcal{V}^\top \nabla \phi = -\lambda_l \phi\mathcal{R}(\vt{x}, t) ~~ \text{in}~~\dom \times \I\;,
}
where $\mathcal{R}(\vt{x}, t) = - \dotp{(\nabla \mathcal{V})\frac{\nabla \phi}{\norm{\nabla \phi}}}{\frac{\nabla \phi}{\norm{\nabla \phi}}}$.
This equation combines the level-set equation and Eikonal constraint, freeing the level-set approach from the reinitialization process~\cite{anonymous2024implicit, fricke2023locally, bothe2023mathematical}.

\inparagraph{Spatial Smoothness and Volume Preservation.} To make sure the velocity field models physical movement, we can impose respective regularizers on the velocity field. Two popular regularizers are spatial smoothness $\mathcal{L}_s$~\cite{dupuis1998, anonymous2024implicit} and volume preservation $\mathcal{L}_v$~\cite{eisenberger2018divergencefree, Cosmo2020}:
\eq{\label{eq:smoothness_volumn}
\mathcal{L}_{s} &= \int_{\dom}\norm{ (-\alpha \Delta + \gamma \vt{I}) \V{\vt{x}}, t}_{l^2} \dd \vt{x} , \\
\mathcal{L}_{v} &= \int_{\dom} | \nabla \cdot \V{\vt{x, t}}| \dd \vt{x} \;.
}

\begin{figure*}[t]
    \centering
    \includegraphics[width=.9\linewidth]{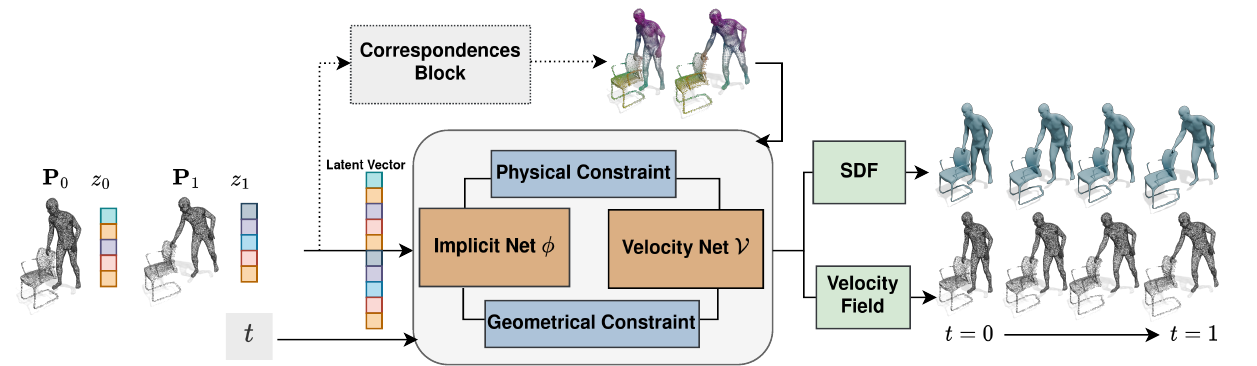}
    \caption{\textbf{Pipeline of 4Deform}: Given a temporal sequence of inputs, we initialize a latent vector to each point cloud. Then the network takes pairs of point clouds $\P_0$ and $\P_1$ (with sparse correspondences), together with the concatenated latent vector $\vt{z}_0$ and $\vt{z}_1$ as input. At training time, we jointly optimize two neural fields: a time-varying implicit representation (Implicit Net $\phi$) and a velocity field (Velocity Net $\mathcal{V}$) with proposed geometric and physical constraints losses. Conditioning on a time stamp $t$, we instantaneously obtain a continuous time-varying signed distance function (SDF), an offset of the input toward the target (velocity field).
}\label{fig:pipeline}
\vspace*{-2mm}
\end{figure*}

\section{... to Neural Implicit Surface Deformation}
\label{sec:method}
Previous implicit or point cloud deformation methods either rely on ground truth correspondences, struggle with large deformations~\cite{anonymous2024implicit}, or are limited to shape pairs~\cite{liu2022learning, Novello2023neural, anonymous2024implicit}, restricting their applicability. To overcome these limitations, we propose a method that first incorporates a corresponding block to obtain sparse correspondences, then handles larger deformations by establishing stronger physical constraints, and extends functionality to \emph{temporal sequence} of point clouds via an AutoDecoder architecture~\cite{Goodfellow2016}. Given point clouds sequence $\{\P_{0}, \P_{1}, \dots, \P_{n},\dots \}$ with an initialized latent vector to each point cloud $\{\vt{z}_0, \vt{z}_1, \dots, \vt{z}_n, \dots\}$, where $\P_{k}=\{\vt{x}_i^k\}_i \subset \mathbb{R}^3$ and $\vt{z}_i \in \mathbb{R}^m$ is a trainable latent vector that is assigned to each input point cloud. We pair the inputs as a source point cloud and a target point cloud (for convenience we label them as $\P_0$ and $\P_1$). We aim to generate physically plausible intermediate stages of all training pairs accordingly. To this end, we propose to model the 4D movements using a time-varying implicit neural field of the form:
\eq{\label{eq:boundary}
\surf{t} &= \{\vt{x} | \phi_{\vt{z}}(\vt{x}, t) =0\}, \text{for}~\vt{z} := \vt{z}_0~\oplus~\vt{z}_1 \;.
}
Each shape $\surf{t}$ in time $t$ is encoded by the zero-crossing of the implicit function $\phi$. Particularly, $\surf{0}$ and $\surf{1}$ should coincide with $\P_0$ and $\P_1$. 
\subsection{Correspondence Block}
Instead of relying on ground-truth correspondences, which are difficult to obtain in real-world settings, our method obtains the correspondences based on the state-of-the-art unsupervised non-rigid 3D shape matching method~\cite{cao2023unsup}. This method is based on the functional map framework~\cite{ovsjanikov2012functional} and follow-up learning-based approaches~\cite{roufosse2019unsupervised,donati2020deep}. The key ingredient of the functional map framework is that, instead of directly finding correspondences in the spatial domain, it encodes the correspondences in the compact spectral domain and thus is robust to large deformation~\cite{cao2023unsup}. More details about functional maps can be found in this lecture note~\cite{ovsjanikov2016computing}. It is worth noting that our method is agnostic to the choice of shape-matching methods. For instance, for specific types of input shapes (e.g. humans), specialized registration methods can also be utilized to obtain more accurate correspondences~\cite{marin2020farm,marin2023geometric,bhatnagar2020loopreg}.    

\subsection{Implicit and Velocity Fields}
To ensure the physically plausible intermediate stage, we model the deformation from the source point cloud to the target point cloud by tracking the point cloud path using the velocity field $\mathcal{V}_{\vt{z}}(\vt{x},t) \in \mathbb{R}^3$. We adopt the velocity field for two reasons, 
\begin{enumerate*}[label=(\roman*), itemjoin=~]
\item The velocity field allows us to control the generated deformation. It is easy to add physical-based constraints directly on velocity to force the intermediate movement to follow certain physical laws. 
\item There is a link from the velocity field to the implicit field as the implicit field can be seen as a macroscopic field. We can perform the material deviation to derive the natural relationship between velocity field $\mathcal{V}_{\vt{z}}(\vt{x},t)$ and implicit field $\phi_{\vt{z}}(\vt{x},t)$. For simplicity, we omit the latent vector $\vt{z}$ in the following equations.
\end{enumerate*}
We follow~\cite{anonymous2024implicit} to use the Modified Level Set equation to link the velocity field and implicit field via level-set loss
\eq{\label{eq:loss_i}
\mathcal{L}_i = \int_{\dom} \norm{\partial _t \phi + \mathcal{V} \cdot \nabla \phi +\lambda_l \phi \mathcal{R}(\vt{x}, t)}_{l^2} \dd \vt{x}\;.
}
Since the surface normal $\vt{n}$ coincides as implicit function gradient $\nabla \phi$, this loss offers \emph{geometry constraint} to Velocity Net, and as the point is moved by Velocity Net, Eq.~\eqref{eq:loss_i} also works as a \emph{physical constraint} from Velocity Net to Implicit Net, as shown in~\cref{fig:pipeline}.
We force the Velocity Net to move the point with known correspondences to the target input position by 
\eq{
\L_m = \int_{\dom^*}\norm{ \vt{x}^0+\int_0^1 \V{\vt{x},\tau}\dd \tau - \vt{x}^1}_{l^2} \dd \vt{x}\;,
}
where $\dom^*$ only contains points with known correspondences. 

\subsection{Physical Deformation Loss}
A key challenge in deforming implicit neural fields with external velocity fields is ensuring physically realistic movement, as commonly used constraints like as-rigid-as-possible (ARAP)~\cite{sorkine2007rigid} are difficult to enforce without a connectivity structure. To address this, we introduce new physical deformation losses on both the implicit and velocity fields to better control movement. These losses do not require connectivity information, thus it can be enforced on unordered input and allow arbitrary resolution of input.

\inparagraph{Distortion Loss.}
In the Eulerian perspective of continuum mechanics, the rate of deformation tensor provides a measure of how the fluid or solid material deforms over time from a fixed reference point in space, excluding rigid body rotations. It measures the rate of stretching, compression, and shear that a material element undergoes as it moves through the flow field~\cite{spencer2004continuum}. The rate of deformation tensor is defined by
\eq{\label{eq:deformation_tensor}
\vt{D} = \frac{1}{2}(\nabla \mathcal{V} + (\nabla \mathcal{V})^\top) \;.
}
The distortion of the particle moved under the velocity $\mathcal{V}$ can be described by the deviatoric form
\eq{\label{eq:distortion}
\mathcal{L}_{d} = \int_{\dom} \norm{\frac{1}{6}\tr(\vt{D})^2-\frac{1}{2} \tr(\vt{D} \cdot \vt{D})^2 }_{F}\dd \vt{x}\;.
}
Eq.~\eqref{eq:distortion} is the complement of the volumetric changes. This term removes the volumetric part, leaving behind the deviatoric (distortional) component. It offers \emph{physical constraint} to both networks during training.

\inparagraph{Stretching Loss.} 
By tracking point cloud movement with a velocity field in our approach, we can establish constraints to control surface stretching along the deformation.
We follow the idea of strain tensor from continuum mechanics~\cite{spencer2004continuum}. Consider deformation happens in infinitesimal time $\Delta t$, the displacement of point $\vt{x}$ is moved to $\vt{x}'$ such that
\eq{\label{eq:displacement}
\vt{x}' = \vt{x} + \V{\vt{x},t}\Delta t\;.
}
Consider a infinitesimal neigbourhood of $\vt{x}'$, denote as $\dd \vt{x}'$, the length of it is given by $ (\dd \vt{s}')^2 = \dd \vt{x}'^\top \dd \vt{x}'$. Similarly, the length of infinitesimal neighborhood of $\vt{x}$ is given by $(\dd\vt{s})^2=\dd\vt{x}^\top\dd\vt{x}$. Together with Eq.~\eqref{eq:displacement}, the stretched length after deformation is given by  
\eq{\label{eq:stretching}
(\dd \vt{s}')^2 - (\dd\vt{s})^2 = \dd\vt{x}^\top (\vt{F}^\top\vt{F}-\vt{I})\dd\vt{x}\;,
}
where $\vt{F} = \partial \vt{x}' / \partial \vt{x} = \vt{I} + \nabla \mathcal{V}$. However, instead of considering the stretch on the neighborhood patch of one surface point, preventing stretching on the tangent plane of a point is what makes a deformation physically realistic~\cite{yang2021geometry}. 
We project $\dd \vt{x}$ in  Eq.~\eqref{eq:stretching} to its tangent space using projection operator $\vt{P}(\vt{x}) = \vt{I} - \vt{n}(\vt{x})^\top\vt{n}(\vt{x})$ to compute the stretching on the tangent plane, where $\vt{n}(\vt{x})$ is the normal vector on point $\vt{x}$. Thus, the stretching on the tangent plane is
\eq{
 (\dd l)^2 = \dd \vt{x}^\top\vt{P}^\top(\vt{x})(\vt{F}^\top\vt{F}-\vt{I})\vt{P}(\vt{x}) \dd \vt{x} \;.
}
Finally, thanks to the nice properties of the implicit field, we have $\vt{n}(\vt{x}) = \frac{\nabla \phi(\vt{x},t)}{\norm{\nabla \phi(\vt{x},t)}}$. Therefore, for any $\dd \vt{x}$, we constraint the matrix Frobenius norm as 
\eq{\label{eq:stretching_loss}
\mathcal{L}_{st} = \int_{\dom} \norm{\vt{P}^\top(\nabla\mathcal{V}^\top\nabla\mathcal{V} + \nabla\mathcal{V} + \nabla\mathcal{V}^\top) \vt{P} }_{F} \dd \vt{x}
}
where $\vt{P} = \vt{I} - \nabla \phi\nabla \phi^\top$.

\subsection{Training and Inference}
\inparagraph{Training.} Given a temporal sequence of inputs $\{\P_k\}_k$, which may be point clouds or meshes, we start by using our correspondence blocks to obtain the correspondences of each training pair.
Importantly, our method does not require full correspondence for every training point; it only requires a subsample of points. During training, we initialize a trainable latent vector for each shape. We concatenate the latent vectors of each training pair and optimize them using our Implicit Net $\phi$ and Velocity Net $\mathcal{V}$ jointly.  We sample $T+1$ discrete time steps uniformly for $t = \in \{0, 1/T,\dots, 1\}$ to compute the loss at each time step. 
The total loss is 
\eq{\label{eq:loss}
\mathcal{L} = \lambda_i \mathcal{L}_i + \lambda_s \mathcal{L}_s + \lambda_v \mathcal{L}_v + \lambda_{st} \mathcal{L}_{st} + \lambda_{m}\L_{m}\;,
}
where $\lambda_i$, $\lambda_s$, $\lambda_v$, $\lambda_{st}$ and $\lambda_{m}$ are  weights for each loss term. For further details about implementation and training, we refer to supplementary materials.

\inparagraph{Inference.} During inference, we give the optimized latent vector for each trained pair into the Implicit Net $\phi$ 
to generate intermediate shapes at different discrete time steps $t$. Given an initial point cloud, we pass it with the optimized latent vector to the Velocity Net $\mathcal{V}$, producing a sequence of deformed points at each time step. 


\section{Experimental Results}
\label{sec:results}

\inparagraph{Datasets.} We validate our method on a number of data from different shape categories. We considered registered human shapes from \textbf{FAUST}~\cite{bogo2014faust}, non-isometric four-legged animals from \textbf{SMAL}~\cite{zuffi2017smal}, and partial shapes from \textbf{SHREC16}~\cite{shrec16}. The input shapes are roughly aligned and we train our correspondence block on each one of them individually~\cite{cao2023unsup}. These datasets do not include temporal sequences, so we train on all possible pair combinations. We also evaluate our method on real-world scans, using motion sequences scans of clothed humans from \textbf{4D-DRESS}~\cite{wang20244ddress}, and noisy Kinect acquisitions of human-object interactions from \textbf{BeHave}~\cite{bhatnagar22behave}. For both cases, correspondences are obtained by template shape registration. Notably, the obtained correspondences are rough estimations, often imprecise, and thus do not guarantee continuous bijective maps between shapes, \eg., due to garments, occlusions, or noise in the acquisitions.


\inparagraph{Baseline methods.} We compare our method against recent NIR-based methods that solve similar problems. LipMLP~\cite{liu2022learning} encourages smoothness in the pair-wise interpolation by relying on Lipschitz constraints; NISE~\cite{Novello2023neural}, similar to us, relies on the level-set equation and uses pre-defined paths to interpolate between neural implicit surfaces. NFGP~\cite{yang2021geometry} relies on a user-defined set of points as handles, together with rotation and translation parameters. We also consider~\cite{anonymous2024implicit} as the most relevant baseline method. Nevertheless, all these methods are tailored for shape pairs. To this end, to evaluate the performance of our method in the context of shape sequences, we compare our method to LIMP~\cite{Cosmo2020}, a mesh-based approach that constructs a latent space and preserves geometric metrics during interpolations.


\inparagraph{Training time.} 
We train all methods on a commodity GeForce GTX TITAN X GPU. Our method needs approximately \emph{8 to 10 minutes per pair}, \ie, for a sequence with 10 pairs, our method takes roughly 1.5 hours.
The work~\cite{anonymous2024implicit} and LipMLP~\cite{liu2022learning} require similar runtimes as our method when training one pair.  NISE~\cite{Novello2023neural} takes around 2 hours for each pair. LIMP~\cite{Cosmo2020} first needs to downsample each mesh to 2,500 vertices and training time takes around 30 minutes per pair. NFGP~\cite{yang2021geometry} requires training separately for each time step and each step
takes 8 to 10 hours, which makes the training time go up to 40 hours for recovering 5 intermediate shapes.

\inparagraph{Evaluation Protocol.} We compare three main settings. First, we train on a single registered shape pair (S) and test the interpolation quality (\textbf{Pairs S/S}). Second, we consider the case of training on registered shape sequences and test the interpolation quality (\textbf{Seq. S/S}), for which we rely on similar metrics. Finally, we consider training on registered shape pairs but \emph{testing on Real point clouds (R)} (\textbf{Pairs S/R}). As metrics, we consider the standard deviation of surface area ($\text{SA}\sigma = \sqrt{\sum_{t=0}^N (A_t-\bar{A})^2 /N}$, where $A_t$ is the surface area for mesh at time $t$ and $\bar{A}$ is the average surface area over the interpolated meshes), which is expected to be close to $0$ for the isometric cases. When we have access to ground truth for the intermediate frames, we also report the Chamfer Distance (CD) and the Hausdorff Distance (HD) errors of the predictions. For the Pairs S/R setting, we also report the pointwise root-mean-square error (P-RMSE), which indicates the Euclidean distance of deformed mesh vertices to the ground truth mesh vertices. In the case a method is not applicable to a certain setting,
we note that with a cross (\xmark).


\subsection{Isometric Shape Interpolation}
\inparagraph{Quantitative comparison.} We show a quantitative comparison of isometric human shapes from 4D-Dress for the three settings in Table~\ref {tab:comp_4d}. We chose this dataset since the frequency of the scans lets us have ground truth for the intermediate frames, as well as access to real scans. Despite competitors being tailored for the Pairs S/S setting, our method performs on par, with better area preservation. Our method also supports sequences, contrary to the majority of previous methods. LIMP fails to generate intermediate shapes that are faithful
to the ground truth. 
We believe that LIMP's poor performance is caused by its strong data demand that is not fulfilled here.
To prove our generalization,  we also show results on the interpolation of SMAL isometric animals in Table~\ref{tab:comp_animal}. The ground-truth evaluation frames are obtained by interpolation of SMAL pose parameters. Our method outperforms the competitors on all the metrics. We report all qualitative results in supplementary materials.

\begin{figure}[ht]
	\centering
	\includegraphics[width=0.9\linewidth]{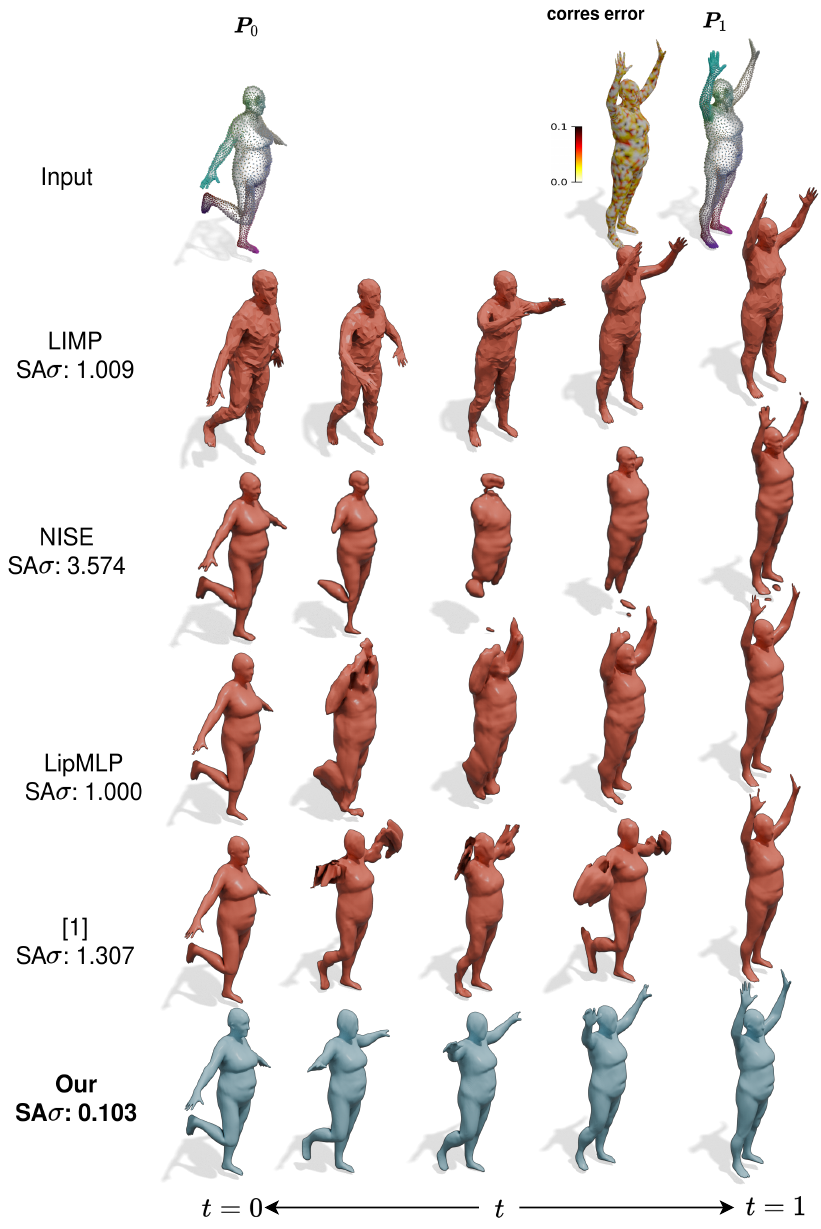}
	\caption{\textbf{Large Deformations.} 4Ddeform handles large deformations better than previous works, providing one order of magnitude less area distortion, even compared to mesh-based ones (LIMP~\cite{Cosmo2020}). In the top row, we visualize the error in the estimated input correspondence.
	\label{fig:large_def}}
 	\vspace{-0.3cm}
\end{figure}

\inparagraph{Large Deformations.} 
A major advantage of our method is that we support large deformations. We show an example of this between two FAUST shapes in Figure~\ref{fig:large_def}. Although isometric, their drastic change of limbs constitutes a challenge for purely extrinsic methods, causing evident artifacts. Our method preserves areas an order of magnitude better than mesh-based methods (LIMP).

\begin{table*}[t]
    \footnotesize
    \addtolength{\tabcolsep}{-3pt}
    \begin{tabular}{lcccccccccccc}
        \specialrule{.1em}{.05em}{.05em} 
        \toprule
         & \multicolumn{3}{c}{Pairs S/S} & &\multicolumn{3}{c}{Seq. S/S} & & \multicolumn{3}{c}{Pairs S/R} \\
        \cmidrule{2-4} \cmidrule{6-8} \cmidrule{10-12}
         &  CD \tiny{$(\times 10^4)\downarrow$} & HD \tiny{$(\times 10^2)\downarrow$} & \SAstd \tiny{$(\times 10)\downarrow$} & 
         &  CD \tiny{$(\times 10^4)\downarrow$} & HD \tiny{$(\times 10^2)\downarrow$} & \SAstd \tiny{$(\times 10)\downarrow$} & 
         &  CD \tiny{$(\times 10^4)\downarrow$} & HD \tiny{$(\times 10^2)\downarrow$} & P-RMSE \tiny{$(\times 10)\downarrow$} \\ 
        \midrule
        LIMP~\cite{Cosmo2020}            &  21.980 & 3.175 & 0.507 & & 136.787& 15.974 & 0.155 & & \xmark & \xmark & \xmark \\
        NFGP~\cite{yang2021geometry}            & 0.272 & \textbf{0.025} & 0.075 & & \xmark  & \xmark & \xmark & & \xmark & \xmark & \xmark \\
        LipMLP~\cite{liu2022learning}          & 14.99 & 2.125 & 1.252 & & \xmark & \xmark & \xmark & &\xmark & \xmark & \xmark \\
        NISE~\cite{Novello2023neural}            & 6.588 & 2.167 & 0.321 & & \xmark & \xmark & \xmark & & \xmark & \xmark & \xmark \\
         \cite{anonymous2024implicit} & 0.279 & 0.047 & 0.023 &&\xmark & \xmark & \xmark && 0.548 & \textbf{0.083} & 0.024  \\
         \midrule
        \textbf{Ours} & \textbf{0.269} & 0.046 & \textbf{0.018} && \textbf{0.327} & \textbf{0.038} & \textbf{0.063} && \textbf{0.390} & 0.101 & \textbf{0.014} \\
        \bottomrule
    \end{tabular}
    \caption{\textbf{Human dataset isometric deformation metrics.}  We evaluate our method on both pairs input (Pairs S/S) and sequence input (Seq. S/S) and compare against previous methods that either only work for temporal sequences or pairs. Additionally, as our method can directly operate on real-world data that is not trained on, we also report the error w.r.t. to real-world results (see~\cref{subsec:application}) on (Seq. S/R). For the methods that cannot directly deform real-world data, thus the Pairs S/R columns are marked as \xmark.
    }
    \label{tab:comp_4d}
    \vspace*{-2mm}
\end{table*}

    
\begin{table}[t]
\footnotesize
\addtolength{\tabcolsep}{-3pt}
\begin{tabular}{lcccc}
\specialrule{.1em}{.05em}{.05em} 
        \toprule
         & \multicolumn{4}{c}{Pairs S/S} \\
         \cmidrule{2-5} 
 & CD \tiny{$(\times 10^3)\downarrow$} & HD \tiny{$(\times 10^2)\downarrow$} & \SAstd \tiny{$(\times 10)\downarrow$} & P-RMSE \tiny{$(\times 10)\downarrow$} \\ 
\cmidrule{2-5}
NFGP~\cite{yang2021geometry} & 0.770 & 0.906 &  0.217 & \xmark\\
LipMLP~\cite{liu2022learning} & 68.452 & 43.327 & 1.192 & \xmark \\
NISE~\cite{Novello2023neural} & 7.223 & 1.237 & 0.771 & \xmark \\
\cite{anonymous2024implicit} & 0.173 & 0.626 & 0.064 & 0.081 \\
\midrule
\textbf{Ours} & \textbf{0.137} & \textbf{0.221} & \textbf{0.062} & \textbf{0.061} \\
\bottomrule
\end{tabular}
\caption{\textbf{Animal dataset isometric deformation metrics.} We show that our method achieves the best results on the SMAL dataset.}
\vspace{-3mm}
\label{tab:comp_animal}
\end{table}

\subsection{Non-isometry and Partiality}
\begin{figure}[ht]
	\centering
 \begin{tikzpicture}[spy using outlines={circle, magnification=2, size=0.55cm, connect spies}]
    \node[anchor=south west,inner sep=0] (image) at (-2, 0){
	\includegraphics[width=\linewidth]{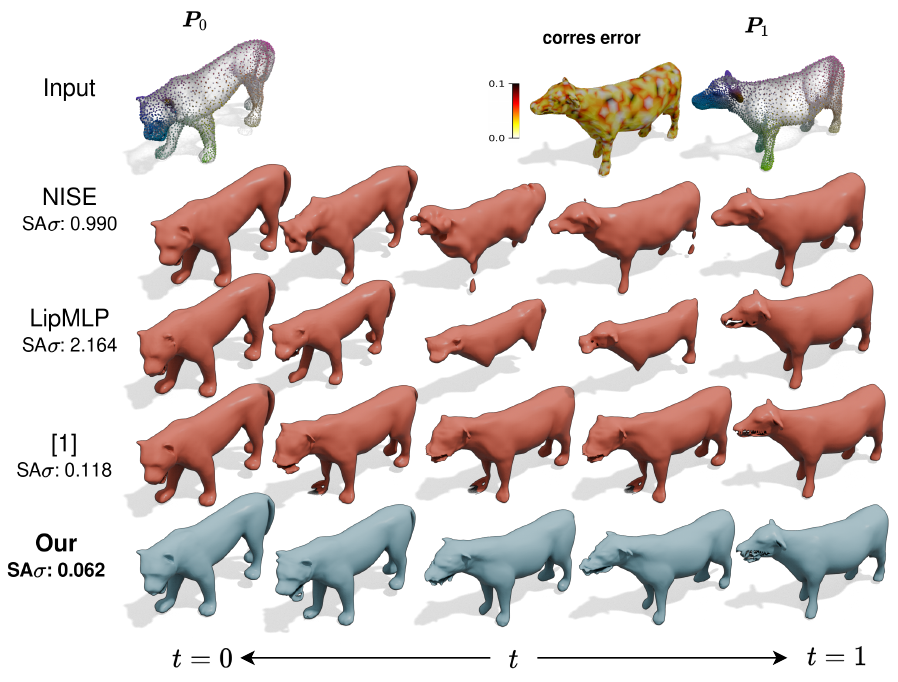}};
 \begin{scope}[x={(image.south west)},y={(image.north east)}]
        \spy[red] on (2.47, 0.68) in node (image) [left] at (3.65,0.59); 
        \spy[red] on (2.44, 1.78) in node (image) [left] at (3.70,1.76);
    \end{scope}
\end{tikzpicture}
	\caption{\textbf{Non-isometric deformation.} We deform two different animals from SMAL, relying on a noisy correspondence (top row). Compared to the previous methods, our method results in plausible deformations, while preserving thin geometric details (e.g., legs).}
	\label{fig:comp_animal}
 	\vspace{-0.3cm}
\end{figure}

\inparagraph{Non-isometry.} A significant challenge is modeling interpolation when the shape metric is drastically changing between frames. This significantly hampers the chances of obtaining reliable correspondence and control over the full-shape geometry. In~\cref{fig:comp_animal}, we show an interpolation between a cougar and a cow from SMAL. As can be seen on top of the image, the correspondence error is quite noisy. As a consequence, our method is the only one that shows consistency also in the thin geometry (e.g., legs). We argue this is a direct contribution of our losses Eq.~\eqref{eq:distortion}  and Eq.~\eqref{eq:stretching_loss}. Due to the lack of ground truth intermediate shapes, we only report \SAstd for each method's results.  

\inparagraph{Partiality.} Finally, an extremely challenging case are partially observed shapes.
Here, an ideal interpolation would provide a smooth interpolation of the overlapping part, while keeping the non-overlapping part as consistent and rigid as possible. We provide an example of a cat from SHREC16 in~\cref{fig:partial_shrec}. Despite the highly imprecise correspondence, we see that our method is the one with the best preservation of the absent area and, overall, the smallest area distortion. For both non-isometric and partial shapes, our method provides more realistic deformations than~\cite{anonymous2024implicit}.

\subsection{Ablation Study}
\begin{figure}[t]
	\centering
 \begin{tikzpicture}
  \node [inner sep=0pt] (image) at (0,0) {
	\includegraphics[width=.9\linewidth]{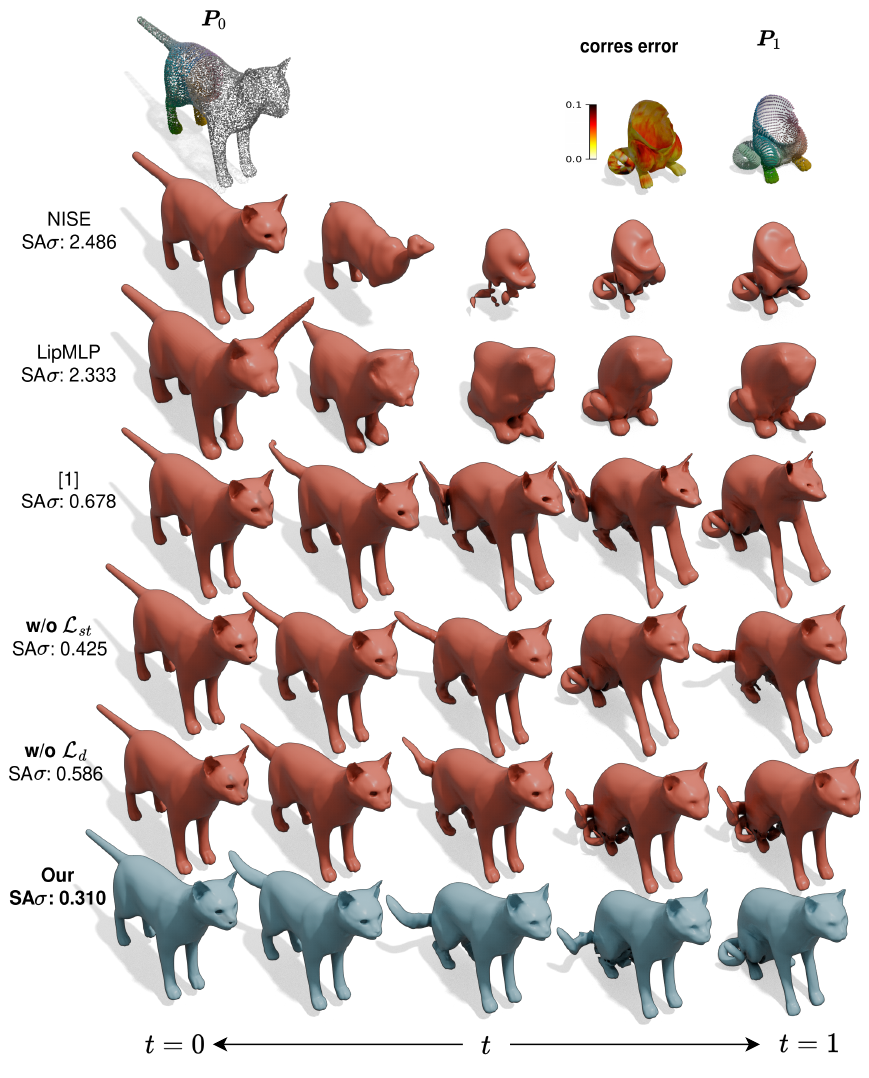}};
    \draw[red, thick] (3.2, -0.2) circle [radius=0.35cm];
    \draw[red, thick] (3.2, -1.45) circle [radius=0.35cm];
    \draw[red, thick] (2.63, -2.4) circle [radius=0.25cm];
    \draw[red, thick] (3.2, -3.9) circle [radius=0.3cm];
    \end{tikzpicture}
	\caption{\textbf{Partial shape deformation.} We consider the case in which one of the input shapes is only partially available while having noisy correspondences (correspondence error visualized in the top row). Other methods often collapse the unseen part or create unreasonable stretches. Similar effects are observed when we remove some of our novel losses. Our method provides plausible interpolations, both for the visible and missing parts.}
	\label{fig:partial_shrec}
 	\vspace{-0.3cm}
\end{figure}

To demonstrate the effectiveness of our distortion loss $\L_{d}$ and stretching loss $\L_{st}$, we perform a quantitative comparison on the 4D-Dress dataset. We report quantitative evaluation in~\cref{tab:ablation_4d}. We highlight that although the losses have a minor influence in the Pair S/S case, they are useful in providing consistency when the network has to capture relations on a wider set of shapes (Seq. S/S); further, they show robustness in the presence of real noise (Pairs S/R). This follows our intuition that such losses serve as regularization, especially in the more challenging cases. This is further highlighted by the qualitative results of partial shapes of~\cref{fig:partial_shrec}.  
\begin{table*}[ht]
    \footnotesize
    \addtolength{\tabcolsep}{-3pt}
    \begin{tabular}{lcccccccccccc}
        \specialrule{.1em}{.05em}{.05em} 
        \toprule
         & \multicolumn{3}{c}{Pairs S/S} & &\multicolumn{3}{c}{Seq. S/S} & & \multicolumn{3}{c}{Pairs S/R} \\
        \cmidrule{2-4} \cmidrule{6-8} \cmidrule{10-12}
        &  CD \tiny{$(\times 10^4)\downarrow$} & HD \tiny{$(\times 10^2)\downarrow$} & \SAstd \tiny{$(\times 10)\downarrow$} & 
         &  CD \tiny{$(\times 10^4)\downarrow$} & HD \tiny{$(\times 10^2)\downarrow$} & \SAstd \tiny{$(\times 10)\downarrow$} & 
         &  CD \tiny{$(\times 10^4)\downarrow$} & HD \tiny{$(\times 10^2)\downarrow$} & P-RMSE \tiny{$(\times 10)\downarrow$} \\ 
        \midrule
         w/o $\mathcal{L}_d$ & \textbf{0.265} & \textbf{0.041} & 0.115 && 0.348  & 0.101  & 0.065 &&
         0.490  & 0.233  & 0.023 \\
        w/o $\mathcal{L}_{st}$  & 0.284 & 0.045 & 0.018  && 0.363 & 0.091 & \textbf{0.062} &&
        0.620  & 0.126  & 0.023 \\
        w/o both & 0.279 & 0.047 & 0.023 && 0.390 & 0.084 & 0.065 &&
        0.548 & \textbf{0.083} & 0.024\\
        \textbf{Ours} & 0.269  & 0.046 & \textbf{0.018} &&  \textbf{0.327} & \textbf{0.038} & 0.063 &&
        \textbf{0.390} & 0.101 & \textbf{0.014}\\
    
\bottomrule
\end{tabular}
\caption{\textbf{Loss ablation.} We ablate our method on pair setting (Paris S/S) and temporal sequences setting (Seq. S/S) and report the quantitative measurements. We also report the error on the real-world mesh CD and HD in Pairs (S/R) columns.
}
\vspace*{-3mm}
\label{tab:ablation_4d}
\end{table*}

\begin{figure*}[th]
	\centering
	\includegraphics[width=.85\textwidth]{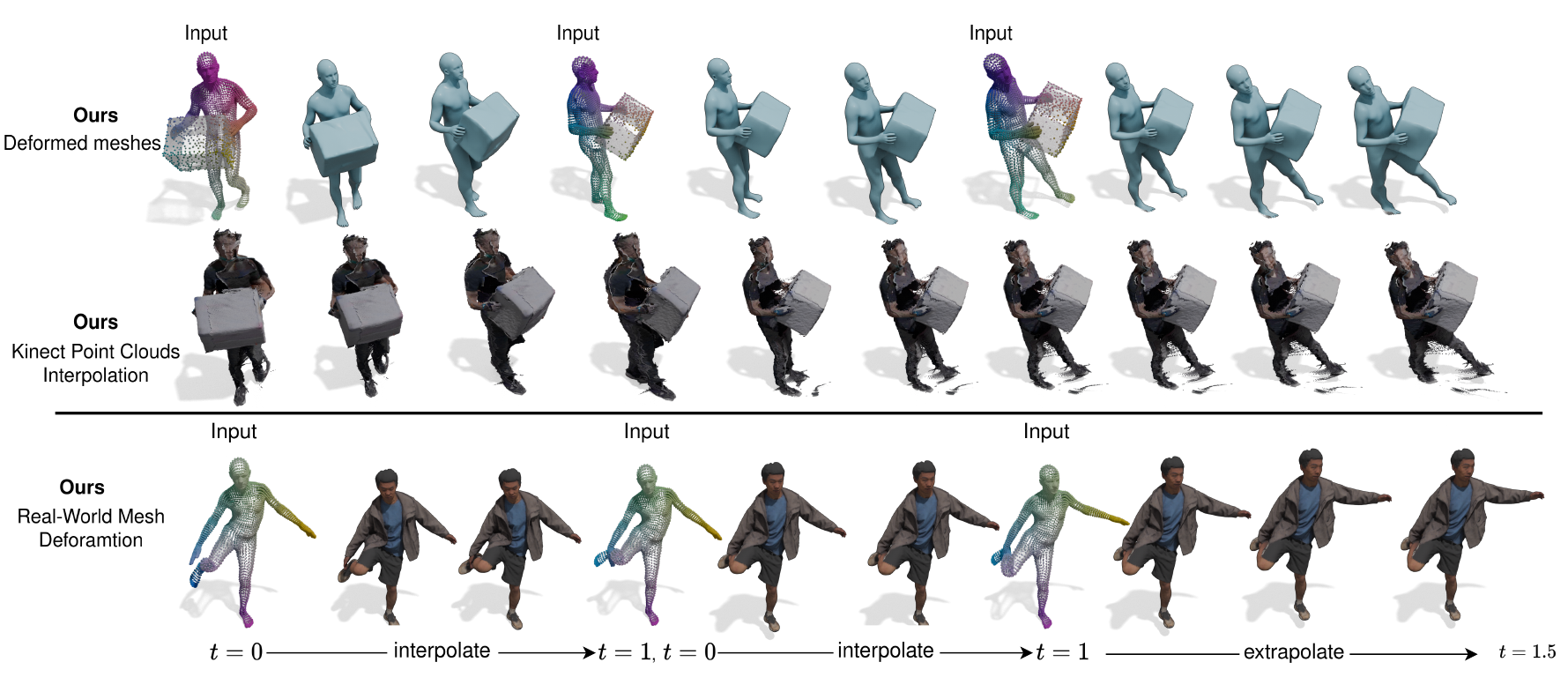}
	\caption{\textbf{Upsampling and extrapolation.} The top shows an example of the BeHave sequence. Starting from a sparse set of keyframes (1fps, colored point clouds), our method lets us interpolate the registration (first row), as well as the real Kinect point clouds (second row) between keyframes at an arbitrary continuous resolution. On the bottom, we show extrapolation on a 4D-Dress sequence. With just a few key frames, we can deform the real point cloud even beyond the final frame, obtaining an estimation of the plausible continuation of the action.}
	\label{fig:app_upsample}
 	\vspace{-0.3cm}
\end{figure*}

\subsection{Applications}\label{subsec:application}
Our method enables a series of new applications, such as the upsampling of real captures and the handling of noisy point clouds. We also show that the learned networks are capable of generalizing and extrapolating in the time domain.

\inparagraph{Temporal Upsampling.}
In many real-world datasets, human movements are recorded by sensors such as RGBD cameras, which provide real-world point clouds of humans and possible  object interactions. However, due to technical constraints or device setup differences, human motion datasets have different frame rates for recording the movement~\cite{bhatnagar22behave, wang20244ddress, GRAB}. 
This is the case for BeHave~\cite{bhatnagar22behave}, where the input Kinect sequences are carefully annotated with significant manual intervention to align SMPL and an object template to the input, resulting in a frame rate of only 1 FPS. We show that our interpolation method can efficiently upsample not only the annotated SMPL data, but also the noisy real-world Kinect point clouds without additional effort. We highlight that our Velocity Net can easily be generalized to untrained real-world point clouds to obtain upsampling sequences. We show the upsampling results in~\cref{fig:app_upsample}

\inparagraph{Real-World Data Deformation.}  Here, we show another application scenario in high-quality meshes.
High-quality data with tens of thousands of vertices, defining a dense, precise point-to-point correspondence is demanding and often unfeasible to be processed. Therefore, it is a common practice to fit a template to such input high-quality data and use it as rough guidance. In~\cref{fig:app_upsample}, we show how, from just a few frames equipped with a rough SMPL alignment,
we can manipulate a 40k vertices real scan, maintaining its structure along all the sequences. We refer to~\cref{tab:comp_4d} as an evaluation of our method's performance both on the real-world and SMPL interpolation.

\inparagraph{Extrapolation for Movement Generation.} Our method lets us obtain dense intermediate frames at arbitrary temporal resolution and allows us to deform the data that is a bit far from the sparse input correspondence. Moreover, the learned physical deformation allows us to generate movements even beyond the considered sequence. Our velocity field can extrapolate outside of the training time domain ($0$ to $1$), while remaining physically plausible, as shown in~\cref{fig:app_upsample}. 

\section{Discussion and Future Work} While our method integrates physical plausibility, certain types of deformations, such as mechanical joints and fluid dynamics, may not yet be fully captured by our model. Future work could incorporate additional physical constraints to address these complexities. Additionally, some applications require separate deformation estimation, as in the case of a human with loose clothing, where the deformations of the body and clothing do not align. We plan to extend our work to handle these cases in future developments.




{
    \small
    \bibliographystyle{ieeenat_fullname}
    \bibliography{main}
}

\clearpage
\setcounter{page}{1}
\setcounter{section}{-1}
\maketitlesupplementary

\renewcommand \thesection{S\arabic{section}}
\renewcommand{\thefigure}{S.\arabic{figure}}
\renewcommand{\thetable}{S.\arabic{table}}
\renewcommand{\theequation}{s.\arabic{equation}}

\section{Contents of Supplementary Materials}
The supplementary zip folder contains the following:
\begin{itemize}
    \item A PDF document providing additional details on our method, including mathematical formulations, training specifics, and extended visualizations of our results.
    \item A video folder showcasing animations of our results. 
    \item A link to our project page \url{https://4deform.github.io/}.
\end{itemize}

\section{Losses Derivations}
\label{sec:math}
\inparagraph{Distortion Loss.} If one breaks down the rate of deformation tensor in~\cref{eq:deformation_tensor}, $\vt{D}$ it is the symmetric part of the velocity gradient $\nabla \mathcal{V}$ plus its transpose. It is called the rate of deformation tensor which gives the rate of stretching of elements. Since $\mathcal{V}:\mathbb{R}^3\to\mathbb{R}^3$, $\vt{D}$ is a $3\times3$ matrix, it is also related to stress tensor in continuum mechanics. We adopt the second invariants of the deviatoric stress tensor~\cite{Irgens2008}
\eq{
J_2 &= \frac{1}{3}\tr(\vt{D})^2 - \frac{1}{2}\big(\tr(\vt{D})^2 - \tr(\vt{D}\cdot \vt{D})\big) \\
 &= \frac{1}{6}\tr(\vt{D}\cdot \vt{D}) - \frac{1}{2}\tr(\vt{D})^2 \;.
}
The second invariant equal to zero implies that there is no shape-changing (distortional) component in the deformation or stress. In this case, all principal stresses or strains are equal, leading to a purely hydrostatic (isotropic) stress or strain state~\cite{timoshenko1956strength, dieter1976mechanical}.

\inparagraph{Stretching Loss} In fact, the term is related to the (right) Cauchy strain tensor and also related to distortion loss. As in~\cref{eq:stretching}, the deformation term $\vt{F}^\top\vt{F} := \vt{C}$ is called the Cauchy strain tensor~\cite{kaye1998definition}. The term $\vt{F}^\top\vt{F} - \vt{I} := \vt{E}$ is called Green-Lagrange strain tensor and used to evaluate how much a given displacement differs locally from a rigid body displacement. Write it in gradient tensor, \ie, $\nabla \mathcal{V}$, we have 
\eq{
\vt{E} = \frac{1}{2}\big( (\nabla\mathcal{V})^\top + \nabla \mathcal{V} + (\nabla\mathcal{V})^\top(\nabla\mathcal{V}) \big)\;.
}
Therefore,~\cref{eq:stretching_loss} can be seen as projecting the rigid displacement to the tangent space of point $\vt{x}$. Even from a different perspective, our formulation coincides with the stretching loss in NFGP~\cite{yang2021geometry}. Different is we have an explicit formulation of deformation operator $\vt{F}$.

\inparagraph{Normal Deformation} Even though our method does not require an oriented point cloud as input. If normal information is available from the given point clouds, one could utilize the natural property of implicit representation to add normal deformation constraints.  We follow the projection from our stretching loss, for any vector $\vt{t}_1$ and $\vt{t}_2$ in the tangent space of point $\vt{x}$ with normal $\vt{n}$, then we have $\vt{n}(\vt{x})\cdot \vt{t}_1 = 0 $ and $\vt{n}(\vt{x}) \cdot \vt{t}_2 = 0 $. The deformation transform $\vt{t}_1$ to $\vt{t}_1' = \vt{F}\vt{t}$, and $\vt{t}_2' = \vt{F}\vt{t}_2$ the $\vt{F}$ is the same as in~\cref{eq:stretching}. Therefore, $\vt{t}_1'$ and $\vt{t}_2'$ lie in the tangent space of the deformed surface point $\vt{x}'$, thus, the normal in $\vt{x}'$, denoted as $\vt{n}'$ should be perpendicular to $\vt{t}_1'$ and $\vt{t}_2'$, that is,
\eq{
\vt{n}'\cdot\vt{t}_1' = 0, ~~\vt{n}'\cdot\vt{t}_2' = 0\;.
}
Then we have 
\eq{
\vt{n}'\cdot \vt{F}\vt{t}_1 = 0, ~~\vt{n}'\cdot \vt{F}\vt{t}_2 = 0\;.
}
This implies 
\eq{
\vt{F}^\top\vt{n}' = \lambda\vt{n}\;. 
}
We normalized it and get the Normal Deformation Loss as  
\eq{
\L_n = \int_{\dom} \norm{\vt{n}_t - \frac{\vt{F}^\top \vt{n}_{t+1}}{\norm{\vt{F}^\top \vt{n}_{t+1}}} }_{l^2} \dd \vt{x}\;.
}



\section{Training Details}
\begin{figure}[h]
    \centering
    \includegraphics[width=0.8\linewidth]{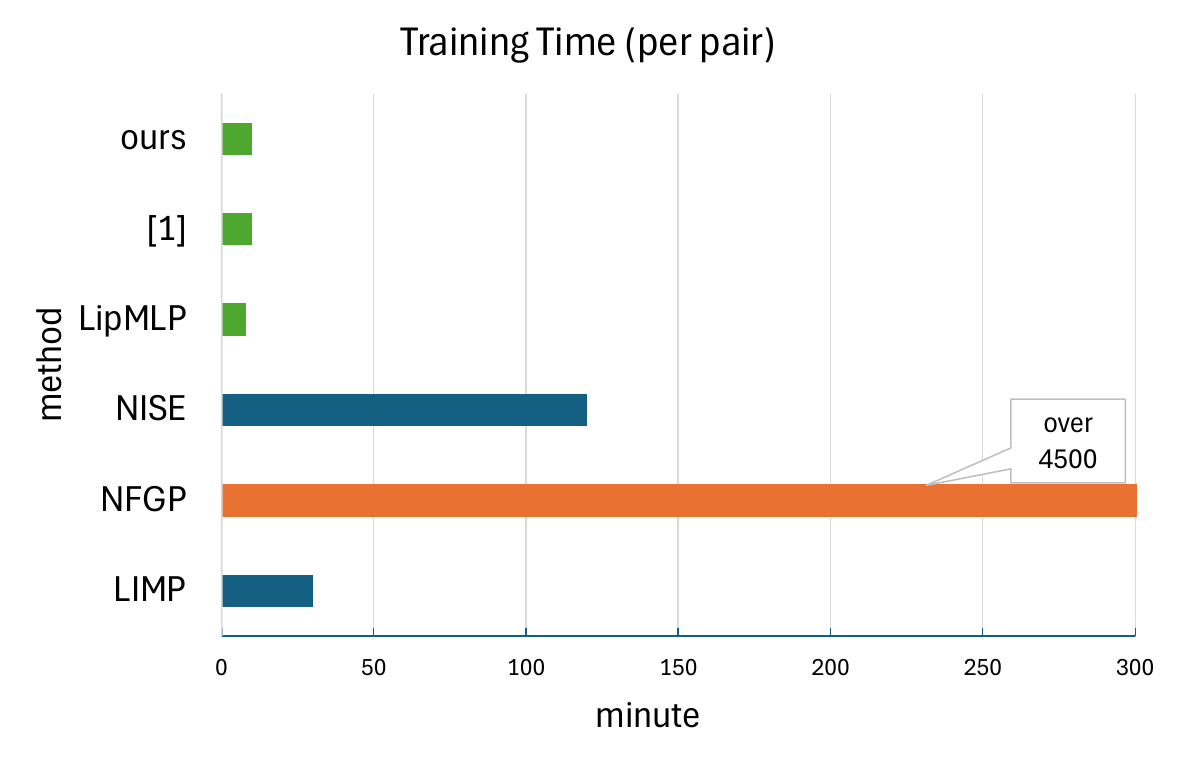}
    \caption{\textbf{Training time visualization.} We plot the rough training time with comparison methods to show the efficiency of our methods.}
    \label{fig:training_time}
\end{figure}

In this section, we summarize the training efficiency of our method and the comparison methods. We plot the average training time (per pair) in~\cref{fig:training_time}. LipMLP~\cite{liu2022learning} trains the fastest as they do not have discrete time steps during training. Our method trains as fast as~\cite{anonymous2024implicit} per pair. However, our method can directly train on temporal sequences without manually switching training pairs. In addition to that, NFGP~\cite{yang2021geometry} requires more than 75 hours to train a 5-step interpolation and LIMP~\cite{Cosmo2020} trains only on meshes with $2,500$ vertices and takes longer than our methods.

\inparagraph{LIMP Training Protocol.} LIMP \cite{Cosmo2020} learns a latent space of meshes and constructs an interpolation constrained under geometric properties. This method supports both isometric and non-isometric deformations. However, the input meshes are required to be in \emph{pointwise correspondence} and \emph{labeled based on stylistic classes}. Additionally, a pre-processing step is needed on the input meshes to reduce the number of vertices to $2500$ and this step is done using iterative edge collapse~\cite{garland1997surface}. The model supports sequence training and training for 20,000 epochs takes about
30-40 minutes for pair training. 

\inparagraph{NISE Training Protocol.} NISE~\cite{Novello2023neural} is a method that learns both isometric and non-isometric deformations between two input meshes. 
It relies on a pair of pre-trained SDF networks to linearly interpolate neural implicit surfaces, which form the foundation for modeling the deformation. In the paper of NISE~\cite{Novello2023neural}, the author mentioned that the method can interpolate along a pre-defined linear path as well. However, this path needs to be defined per point and it can only interpolate linearly according to the Euclidean coordinates of the points. The method can only be trained on mesh pairs, and training each pair, including pre-training the SDF network to fit the input, requires approximately 4 hours for 20,000 epochs. Excluding the pre-training time is approximately 2 hours per pair.

\inparagraph{NFGP Training Protocol.} Training NFGP~\cite{yang2021geometry} requires first training an SDF network that fits the implicit field on the input shapes, which takes about 2 hours for 100 epochs. After that, a set of points is defined per deformation step as handles, along with the necessary rotation and translation parameters to transform these handle points into target points. To be able to use NFGP~\cite{yang2021geometry} as a time-dependent interpolation network that generates $t$ intermediate shapes, one needs to train the network $t$ times and decide how the gradual deformation at each time step should appear. Therefore, the process of defining handle points requires a thorough understanding of how to set rotations and translations to obtain physically plausible interpolation. Moreover, visualization is essential for selecting handles and targets from the vertices of the meshes reconstructed from their SDF network. The training for 500 epochs per deformation time step takes 8 hours. Thus, 50 hours — including the training for the implicit network — are required for deformation with 5 time steps.

\section{Visualizations of Quantitative Evaluated Sequences}
\begin{figure}[t]
    \centering
    \includegraphics[width=\linewidth]{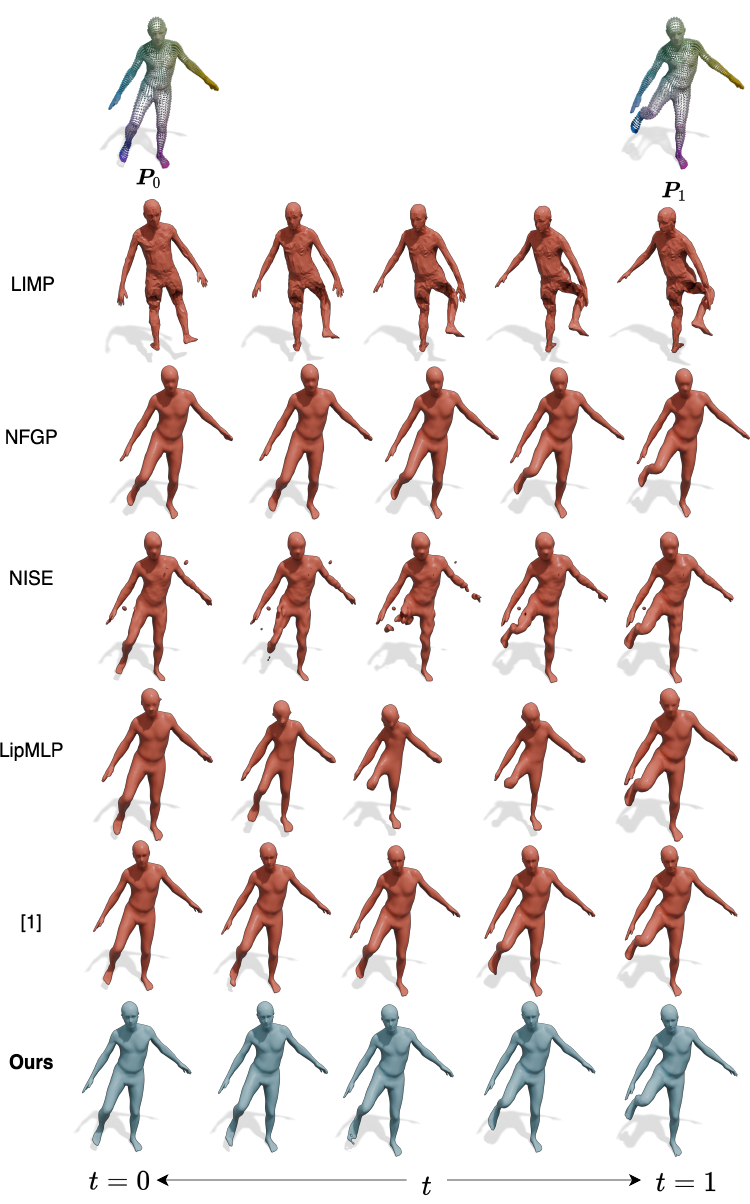}
    \caption{\textbf{Visual results on human isometric deformation.} We show the visualization of our interpolated meshes on~\textbf{4D-Dress}~\cite{wang20244ddress}. LIMP~\cite{Cosmo2020} can recover reasonable movement, however, it turns the leg in the wrong direction.}
    \label{fig:4d_dress}
\end{figure}

\begin{figure}[t]
    \centering
    \includegraphics[width=\linewidth]{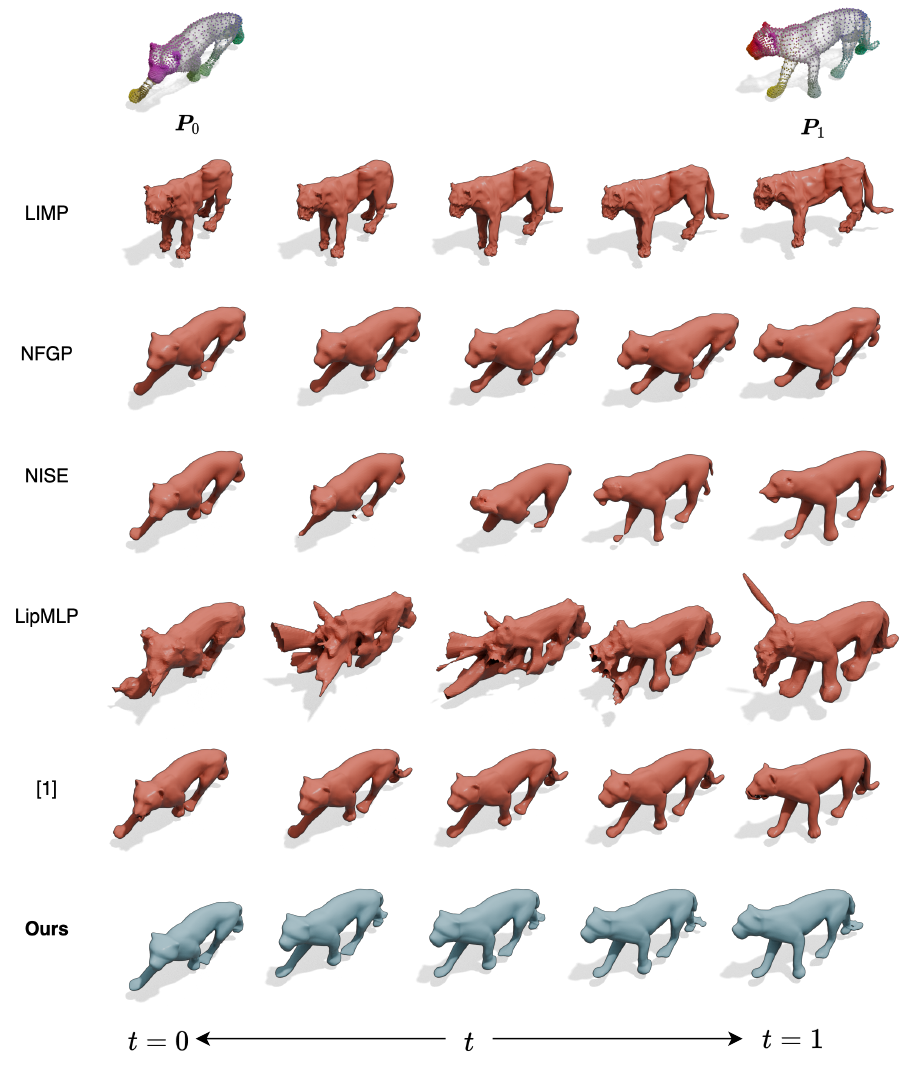}
    \caption{\textbf{Visual results on non-human object deformation.} We show the visualization results for an animal data in~\textbf{SMAL}~\cite{zuffi2017smal}. LIMP~\cite{Cosmo2020} can only handle $2,500$ vertices, thus the interpolated mesh is low-quality.}
    \label{fig:lion}
\end{figure}

In this section, we show the visual results of~\cref{tab:comp_4d} on \textbf{4D-Dress}~\cite{wang20244ddress} in~\cref{fig:4d_dress}. And the visual results of~\cref{tab:comp_animal} on~\textbf{SMAL}~\cite{zuffi2017smal} dataset in~\cref{fig:lion}, to show the deformation of non-human objects. 

\section{More Visualization}
\begin{figure*}[h]
    \centering
    \includegraphics[width=\linewidth]{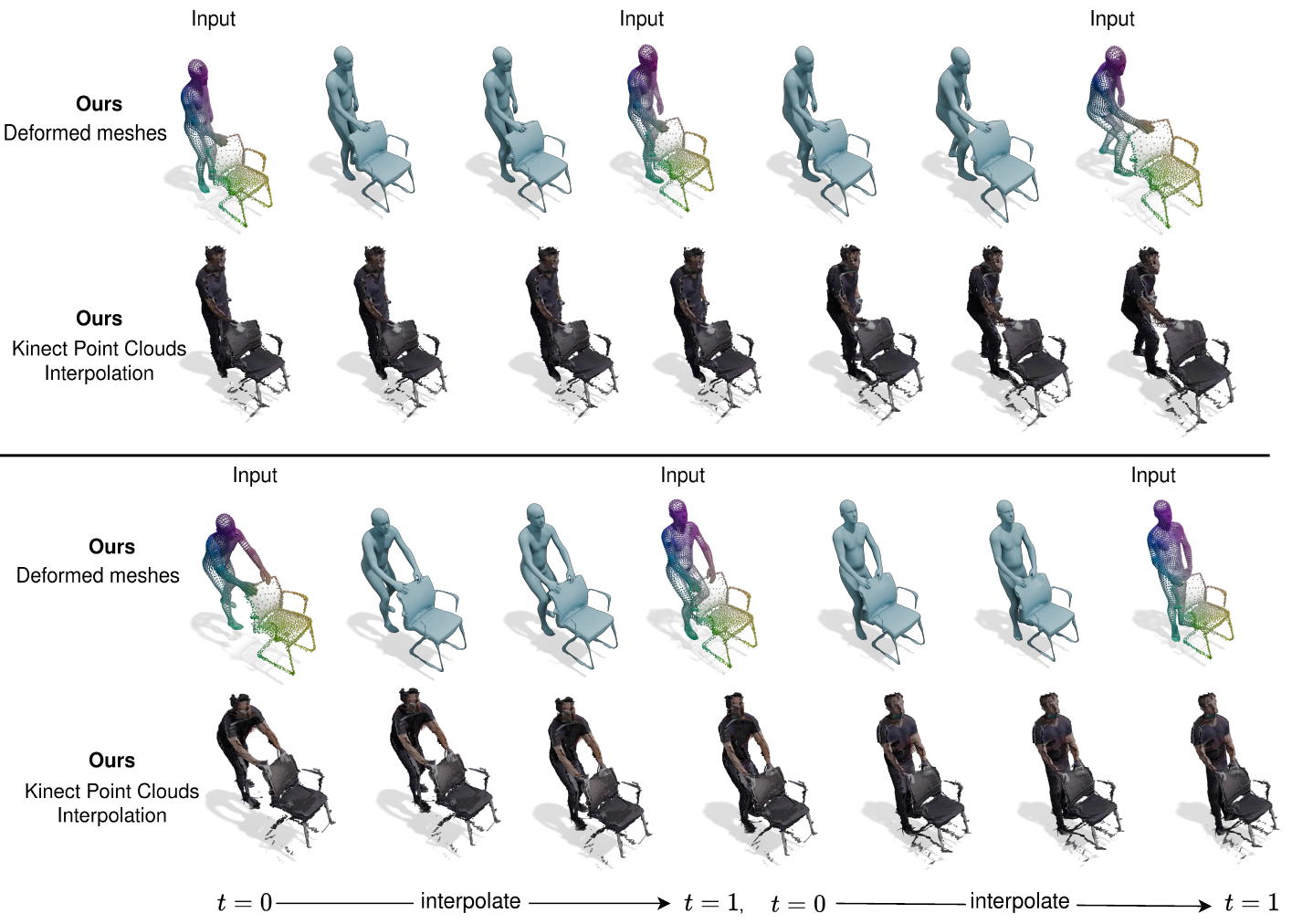}
    \caption{\textbf{Upsampling on real-world data.} We show examples of the \textbf{BeHave}~\cite{bhatnagar22behave} sequence. Starting from a sparse set of keyframes (1fps, colored point clouds), our method lets us interpolate the registration (first row), as well as the real Kinect point clouds (second row) between keyframes at an arbitrary continuous resolution.}
    \label{fig:chairblack}
\end{figure*}

\begin{figure*}[h]
    \centering
    \includegraphics[width=\linewidth]{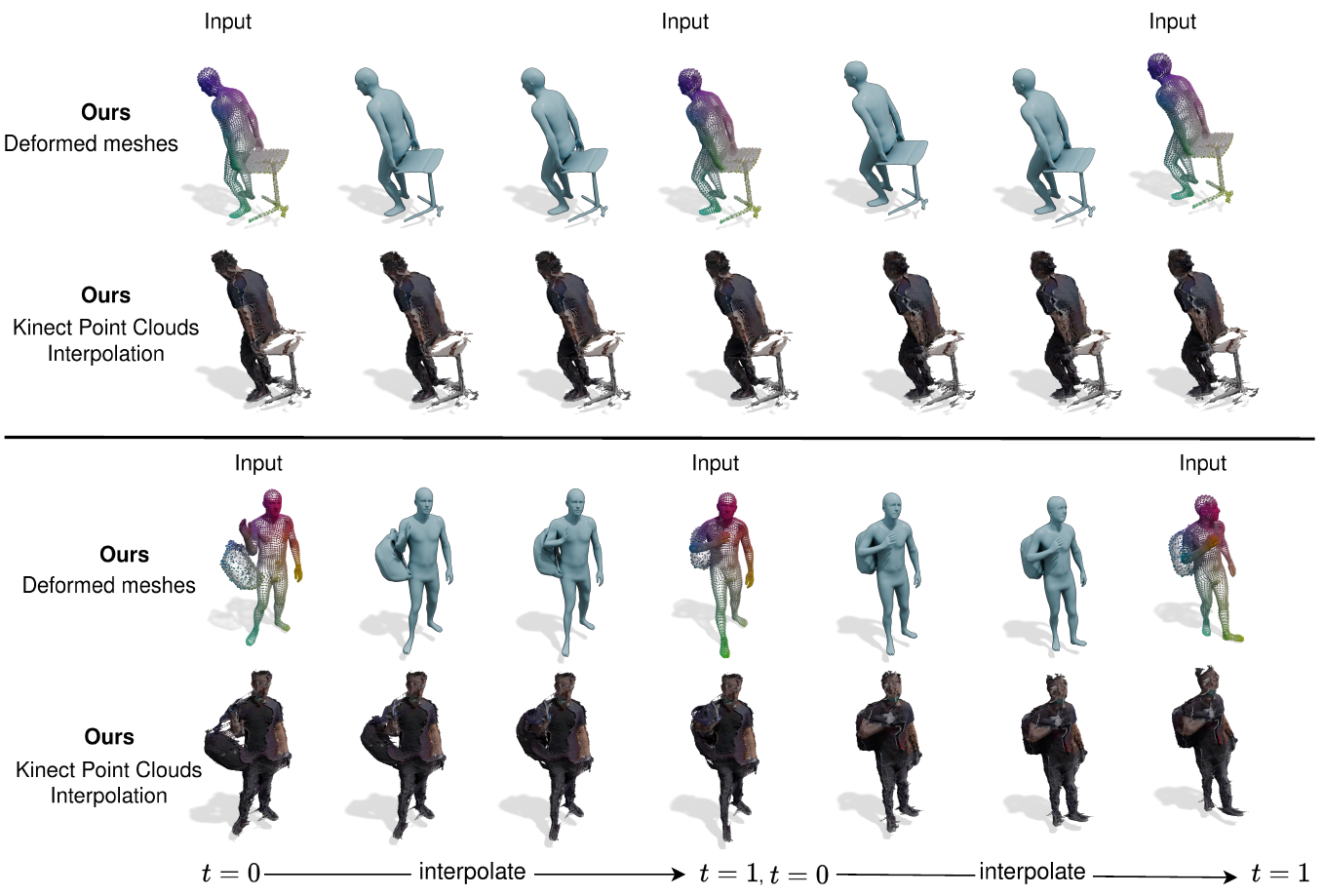}
    \caption{\textbf{Upsampling on real-world data.} We show examples of the \textbf{BeHave}~\cite{bhatnagar22behave} sequence. Starting from a sparse set of keyframes (1fps, colored point clouds), our method lets us interpolate the registration (first row), as well as the real Kinect point clouds (second row) between keyframes at an arbitrary continuous resolution.}
    \label{fig:backpack}
\end{figure*}

\begin{figure*}[h]
    \centering
    \includegraphics[width=\linewidth]{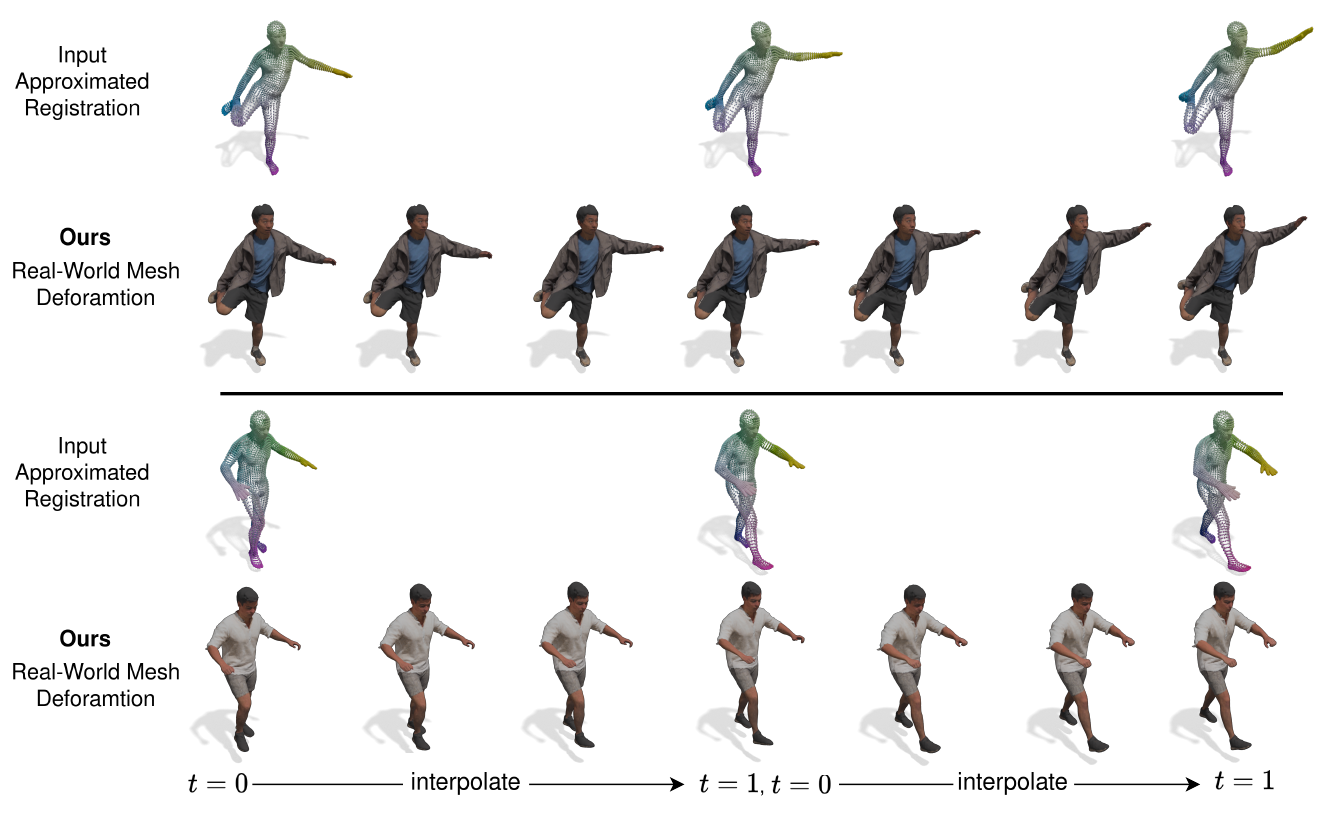}
    \caption{\textbf{Deformation on real-world mesh.} We examples of the \textbf{4D-Dress}~\cite{wang20244ddress} sequence. Starting from a sparse set of approximated registration of SMPL model~\cite{smpl}, our method lets us interpolate the real-world, high-resolution meshes (second row, around $40,000$ vertices) between keyframes.}
    \label{fig:dress4d_supp}
\end{figure*}

We show more visualization results of our method on real-world datasets. We show more sequences from \textbf{BeHave}~\cite{bhatnagar22behave} in~\cref{fig:chairblack} and \cref{fig:backpack}. We also show more visualization of high-resolution real-world mesh interpolation on~\textbf{4D-Dress}~\cite{wang20244ddress} in~\cref{fig:dress4d_supp}.

\section{Upsampling Video}
We use our method to upsample sequences in~\textbf{BeHave}~\cite{bhatnagar22behave} to $30$FPS and render video for it. Please visit our anonymous project page \url{https://4deform.github.io/}. 
%



\end{document}